\documentclass[runningheads]{llncs}
\usepackage[T1]{fontenc}
\usepackage{booktabs}
\usepackage{graphicx}
\usepackage{amsmath}
\usepackage{subcaption}
\usepackage{xcolor}
\usepackage[most]{tcolorbox}
\usepackage{multirow}

\usepackage{pgfplots}
\pgfplotsset{compat=1.18}
\usepgfplotslibrary{groupplots}
\usepackage{url}
\usepackage{siunitx}
\begin{document}
\title{TITAN: Graph-Executable Reasoning for \\Cyber Threat Intelligence}


\author{Marco Simoni\inst{1,2} \and
Aleksandar Fontanta\inst{2,3} \and
Andrea Saracino\inst{3} \and
Paolo Mori\inst{2}}

\authorrunning{M. Simoni et al.}

\institute{Sapienza Università di Roma, Piazzale Aldo Moro 5, 00185 Roma, Italy \and
Institute of Informatics and Telematics (IIT), National Research Council of Italy (CNR), Via G. Moruzzi 1, 56124 Pisa, Italy \and
Department of Excellence in Robotics and AI, TeCIP Institute, Scuola Superiore Sant'Anna, Piazza Martiri della Libertà 33, 56127 Pisa, Italy}

\maketitle              % typeset the header of the contribution
\begin{abstract}

\textit{TITAN} (\textit{Threat Intelligence Through Automated Navigation}) is a framework that connects natural-language cyber–threat queries with \emph{executable reasoning} over a structured knowledge graph.  
It integrates a \textit{path-planner model}, which predicts logical relation chains from text, and a \textit{graph executor} that traverses the TITAN Ontology to retrieve factual answers and supporting evidence.  
Unlike traditional retrieval systems, TITAN operates on a typed, bidirectional graph derived from MITRE ATT\&CK, allowing reasoning to move clearly and reversibly between threats, behaviors, and defenses. To support training and evaluation, we introduce the \textit{TITAN Dataset}, a corpus of 88{,}209 examples (Train: 74{,}258; Test: 13{,}951) pairing natural-language questions with executable reasoning paths and step-by-step Chain-of-Thought explanations.  
Empirical evaluations show that TITAN enables models to generate syntactically valid and semantically coherent reasoning paths that can be deterministically executed on the underlying graph. The code of TITAN is available at this link\footnote{\url{https://github.com/cti-graph-reasoner/TITAN}}.

\end{abstract}

\section{Introduction}

Cyber Threat Intelligence (CTI) relies on the ability to analyze and reason over heterogeneous data describing threat actors, malware behaviors, and defensive measures.  
Recent studies applied Retrieval-Augmented Generation (RAG) to CTI~\cite{fieblinger2024actionable,simoni2025morse}, improving factuality but struggling with \textit{multi-hop} queries that require reasoning across multiple entities~\cite{cyb_with_llm_rag}.  
Most existing systems still rely on entity retrieval or rule-based querying~\cite{MATRIX}, lacking the capacity to generate and execute relational reasoning chains.

To address these limitations, we present \textit{TITAN} (\textit{Threat Intelligence Through Automated Navigation}), a framework designed to autonomously traverse a CTI knowledge graph in order to retrieve the information that accurately answers a given CTI question. 
 It consists of two core components: a Large Language Model (LLM) called \textit{path planner}, inspired by~\cite{rog}, and a \textit{graph executor}.  
Given a natural-language query, the planner predicts logical relational paths, along with the starting entities, that capture the reasoning process needed to answer the question.  
The executor then runs these paths on a modified version of the MITRE\footnote{\url{https://github.com/mitre/cti}} knowledge graph (KG), traversing the corresponding relations to retrieve the final entities that satisfy the query. We modified the KG to follow what we called \textit{TITAN Ontology}~\cite{liu2020stix}, which defines a directed, typed, and bidirectional CTI knowledge schema.  
From this ontology, we automatically built the \textit{TITAN Dataset}, a large collection of natural-language questions paired with their\textit{ Chain-of-Thought} (CoT)~\cite{wei2022chain} explanations and executable relation-based paths. Experiments conducted on this dataset show that the CoT-based model consistently outperforms a direct, non-reasoning (\textit{NoCoT}) baseline, with the largest improvements observed on long and multihop paths. In the paper, we present the related work (Section~\ref{sec:related}), describe the TITAN framework (Section~\ref{sec:framework}), report the experiments (Section~\ref{sec:experiments}), and ends with conclusions and future work (Section~\ref{sec:conclusions}).

\section{Related Work}
\label{sec:related}
Knowledge graphs organize structured threat data for cybersecurity analysis. Frameworks like MITRE ATT\&CK~\cite{strom2018mitre} and CyGraph~\cite{noel2016cygraph} model attacker behaviors through relational structures, but they mainly represent static links and lack mechanisms for dynamic reasoning. To address this, methods such as A4CKGE~\cite{uncovering-attacks-with-KG} extend these frameworks with embedding-based link prediction, though their inference process remains largely opaque. Recent Cyber Threat Intelligence (CTI) automation leverages large language models to extract and correlate threat data~\cite{survey-cyber-threat-detection}, while RAG-based systems improve factual grounding in CTI~\cite{fieblinger2024actionable,simoni2025morse}. However, they struggle with multi-hop reasoning and compositional inference across entities~\cite{cyb_with_llm_rag}, often relying on rule-based or retrieval-based methods~\cite{MATRIX} and offering limited transparency~\cite{XAI-for-cybersecurity}. Although language models can perform relational reasoning on graphs~\cite{GRaph-LLM,IBM-KG-path}, their application to cybersecurity remains limited. \textit{TITAN} addresses these gaps by having the \textit{planner} identify the subject and target entities from the user question, then generate reasoning paths that would logically satisfy the request. The paths are then traversed using the \textit{graph executor}, which at that point is completely deterministic and traceable. 
\begin{figure}[t]
    \centering
    \includegraphics[width=1\linewidth]{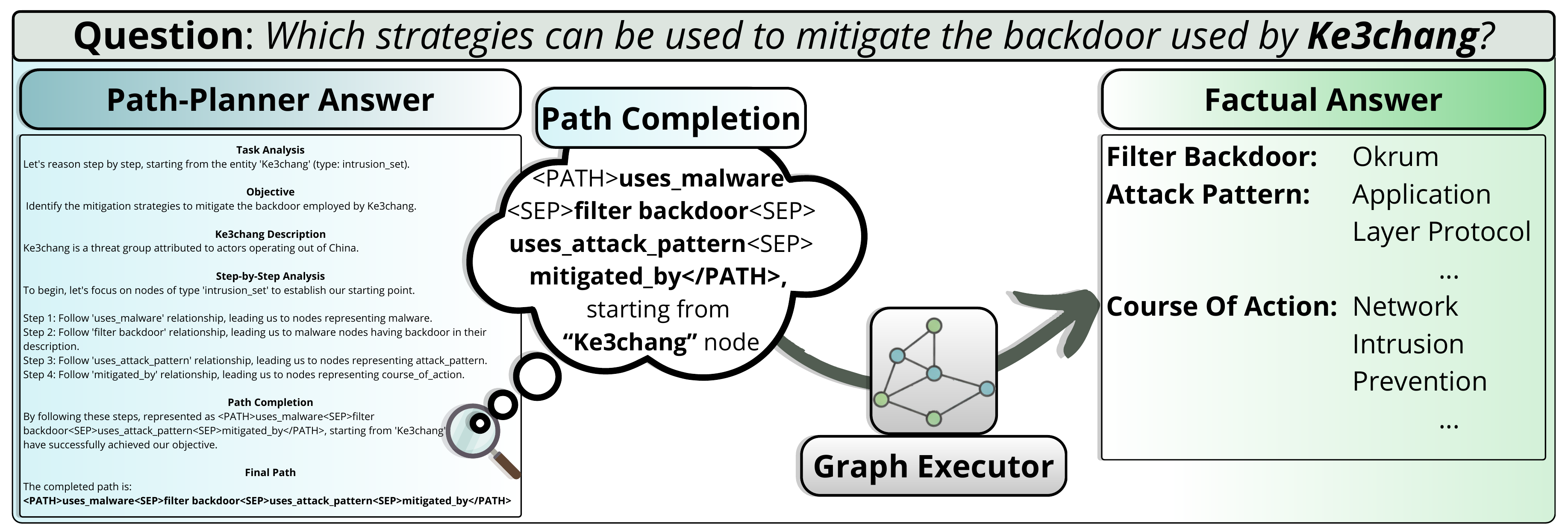}
    \caption{TITAN Framework}
    \label{fig:titan_framework}
\end{figure}

\section{TITAN Framework}
\label{sec:framework}

As shown in figure~\ref{fig:titan_framework}, TITAN operates through two main components: a LLM called \textit{path-planner} and a \textit{graph executor}.  
Given a natural-language query, the model predicts, along with the starting nodes, a logical relational paths that represent the reasoning process needed to answer it. These generated paths are then executed on the TITAN KG (2,350 nodes, 48,795 edges) where the executor traverses the corresponding relations, and retrieves the entities that satisfy the query. Importantly, the entities retrieved by executing the last traversed path are the ones which are supposed to satisfy the request of the query.
The TITAN KG contains all the nodes that are in the MITRE KG but the edges respect the conventions of the \textit{TITAN Ontology}. In the following, we explain the ontology structure and the \textit{TITAN Dataset} derived through it.

\begin{figure}[t]
    \centering
    \includegraphics[width=0.8\textwidth]{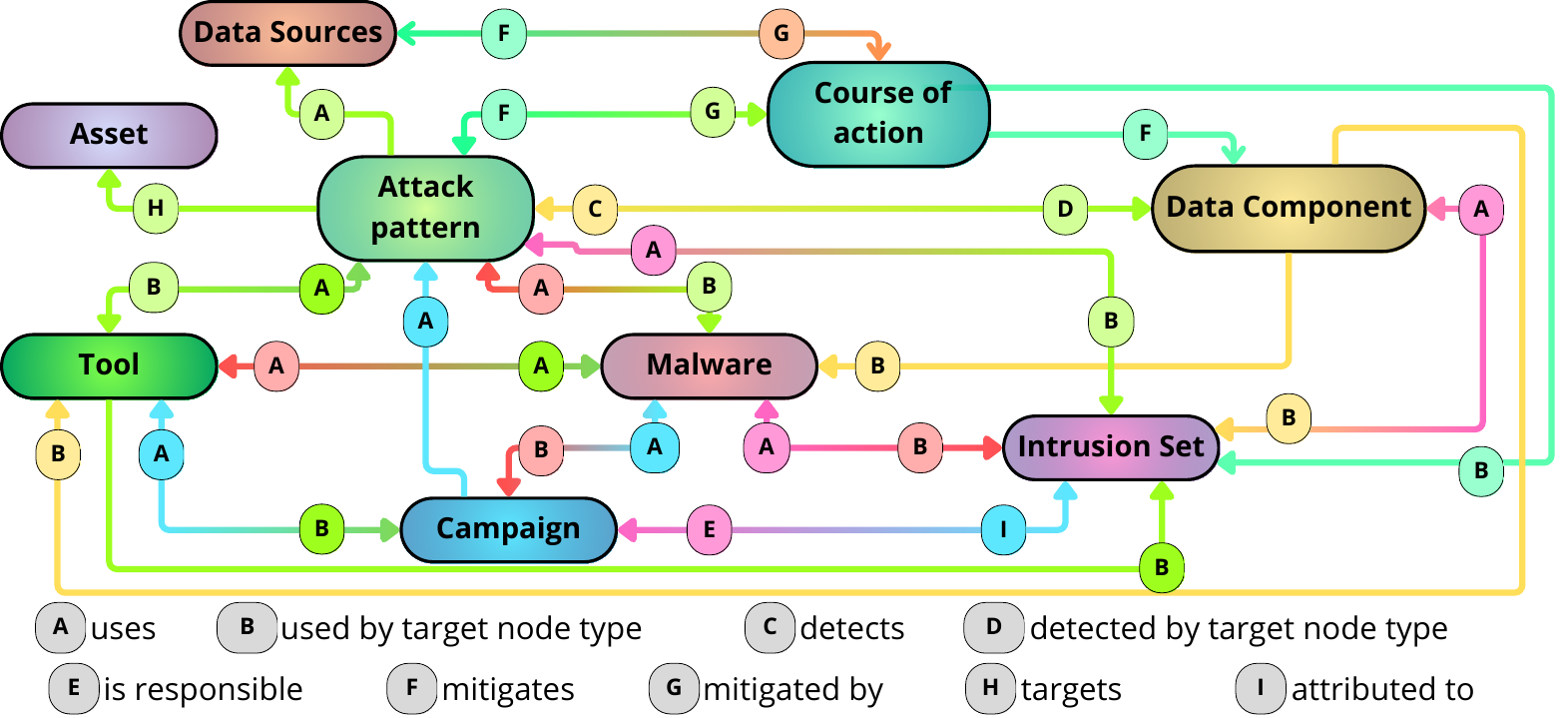}
    \caption{TITAN Ontology Structure}
    \label{fig:ontology}
\end{figure}
\paragraph{\textbf{TITAN Ontology.}}
\label{sec:titan_ontology}

As shown in Figure~\ref{fig:ontology}, the ontology is composed of several entity, each representing a distinct abstraction level within CTI.  
\textit{Attack Patterns} (883 nodes in TITAN KG) define the concrete techniques malware employ to compromise targets.  
\textit{Courses of Action} (318 nodes in TITAN KG) describe defensive strategies that can mitigate or respond to these techniques.  
\textit{Malware} (732 nodes in TITAN KG) nodes model malicious software families or samples, whereas \textit{Tools} (89 nodes in TITAN KG) represent legitimate applications repurposed for malicious purposes.  
At a higher level, \textit{Campaigns} (31 nodes in TITAN KG) aggregate adversarial operations over time, and \textit{Intrusion Sets} (168 nodes in TITAN KG) represent the organized threat actors coordinating them.  
\textit{Data Components} (122 nodes in TITAN KG) and \textit{Data Sources} (43 nodes in TITAN KG) provide the observational layer, defining the evidence and telemetry used to detect malicious activities.  
Finally, \textit{Assets} (14 nodes in TITAN KG) specify the OS and assets  targeted by techniques. We designed all relations in TITAN to be both \textit{typed} and \textit{bidirectional}.  
A \textit{typed} relation explicitly encodes the type of its target node within the relation label, for example distinguishing between \texttt{uses\_attack\_pattern} and \texttt{uses\_malware}.  
This design removes ambiguity among edges that share the same verb (e.g., \textit{uses}) but connect to different entity types. Without typed relationships the graph executor does not know where to move if for example, at a specific node, multiple \textit{uses} relationships connects different target entity types.
\textit{Bidirectional} means that every relation in the graph is represented in both directions, creating a forward and a reverse edge with consistent semantics.
This allows reasoning to move freely between causes and effects, for example, from a malware sample to the techniques it uses, or from a technique back to all malware that employ it.
Concretely,
\textit{malware} $\xrightarrow{\texttt{uses\_attack\_pattern}}$ \textit{attack pattern}
$\quad\Longleftrightarrow\quad$
\textit{attack pattern} $\xrightarrow{\texttt{used\_by\_malware}}$ \textit{malware}.
As a result, the model can start its reasoning from any node.

% \begin{figure}[h!]
%     \centering
%     \includegraphics[width=0.7\textwidth]{TITAN.png}
%     \caption{TITAN Ontology Structure}
%     \label{fig:ontology}
% \end{figure}

\paragraph{\textbf{TITAN Dataset.}}
\label{sec:titan_dataset}

\begin{figure*}[t]
\begin{minipage}{1.01\textwidth}
\scriptsize
% ===== L4+ =====
\begin{tcolorbox}[
    colback=blue!10!white,
    colframe=blue!80!black,
    boxsep=0pt,
    left=1pt, right=1pt, top=1pt, bottom=-0.5pt, % <-- qui
    enhanced jigsaw,
    before skip=0pt, % spazio sopra
    after skip=0.5pt   % spazio sotto
]
\scriptsize
\textbf{L4 - Question}: \textit{What countermeasures can be put in place to prevent the spread of the malware employed by groups responsible for the \texttt{Unitronics Defacement Campaign}?}

% --- BOX GRIGIO INCISO SUI BORDI ---
\begin{tcolorbox}[
    colback=gray!10!white,
    colframe=gray!80!black,
    boxrule=0.5mm,
    enhanced jigsaw,
    boxsep=1pt,
    left=0pt, right=0pt, top=0.5pt, bottom=0pt,
    before skip=0.5pt, after skip=0pt
]
\textbf{Path}: \texttt{attributed\_to → uses\_malware → uses\_attack\_pattern → mitigated\_by}
\end{tcolorbox}

\end{tcolorbox}

% ===== filter =====
\begin{tcolorbox}[
    colback=blue!10!white,
    colframe=blue!80!black,
    boxsep=0pt,
    left=1pt, right=1pt, top=1pt, bottom=-0.5pt, % <-- qui
    enhanced jigsaw,
    before skip=0pt, % spazio sopra
    after skip=0.5pt   % spazio sotto
]
\scriptsize
\textbf{\texttt{filter} - Question}: \textit{Which attacks used by Russian groups target Windows systems?}
\begin{tcolorbox}[
    colback=gray!10!white,
    colframe=gray!80!black,
    boxrule=0.5mm,
    enhanced jigsaw,
    boxsep=1pt,
    left=0pt, right=0pt, top=0.5pt, bottom=0pt,
    before skip=0.5pt, after skip=0pt
]
\textbf{Path}:\texttt{is\_intrusion\_set\_type → filter Russia → uses\_attack\_pattern → targets → filter Windows}
\end{tcolorbox}
\end{tcolorbox}

% ===== exec_common =====
\begin{tcolorbox}[
    colback=blue!10!white,
    colframe=blue!80!black,
    boxsep=0pt,
    left=1pt, right=1pt, top=1pt, bottom=-0.5pt, % <-- qui
    enhanced jigsaw,
    before skip=0pt, % spazio sopra
    after skip=0.5pt   % spazio sotto
]
\scriptsize
\textbf{\texttt{exec\_common} - Question}: \textit{Which techniques are commonly used by \textit{Cannon} and \textit{LitePower}?}
\begin{tcolorbox}[
    colback=gray!10!white,
    colframe=gray!80!black,
    boxrule=0.5mm,
    enhanced jigsaw,
    boxsep=1pt,
    left=0pt, right=0pt, top=0.5pt, bottom=0pt,
    before skip=0.5pt, after skip=0pt
]
\textbf{Path}: \texttt{ is\_malware\_type → select Cannon LitePower → uses\_attack\_pattern → exec\_common}
\end{tcolorbox}
\end{tcolorbox}

% ===== exec_difference =====
\begin{tcolorbox}[
    colback=blue!10!white,
    colframe=blue!80!black,
    boxsep=0pt,
    left=1pt, right=1pt, top=1pt, bottom=-0.5pt, % <-- qui
    enhanced jigsaw,
    before skip=0pt, % spazio sopra
    after skip=0.5pt   % spazio sotto
]
\scriptsize
\textbf{\texttt{exec\_difference} - Question}: \textit{What techniques are used by \textit{Sys10} but not by \textit{MarkiRAT}?}
\begin{tcolorbox}[
    colback=gray!10!white,
    colframe=gray!80!black,
    boxrule=0.5mm,
    enhanced jigsaw,
    boxsep=1pt,
    left=0pt, right=0pt, top=0.5pt, bottom=0pt,
    before skip=0.5pt, after skip=0pt
]
\textbf{Path}: \texttt{is\_malware\_type → select Sys10 MarkiRAT → uses\_attack\_pattern → exec\_difference}
\end{tcolorbox}
\end{tcolorbox}

\end{minipage}
\caption{\small TITAN examples across path depths (L1–L4+) and operators (\texttt{filter}, \texttt{select}, \texttt{exec\_common}, \texttt{exec\_difference}).}
\label{fig:titan_examples_blocks}
\end{figure*}

The TITAN Dataset is automatically generated from the Ontology (Fig. \ref{fig:ontology}), as it enables the identification of all valid reasoning relational paths that connect entities in the graph. Reasoning paths represent logical sequences for answering natural-language questions and they depend only on node types (e.g. malware), not on the specific entities (e.g. Emotet). For instance, the question \textit{“Which strategy mitigates the attack pattern used by the malware Emotet?”} can be satisfied by traversing these reasoning paths:
\texttt{uses\_attack\_pattern → mitigated\_by\_course\_of\_action}. In this case, any malware entity (e.g., \texttt{TrickBot}, \texttt{AgentTesla}) could replace \texttt{Emotet} without changing the paths. For this reason, we manually built more than 700 question templates with typed placeholders (e.g., \texttt{[malware]}) which are automatically filled with entities from the TITAN KG. To increase linguistic diversity in the training set and prevent the planner from learning the template formats, we apply paraphrasing through LLMs as done by~\cite{alajramy2025device}.
At each question template, we associate the reasoning paths to traverse to satisfy the question (e.g. \texttt{<PATH> uses\_attack\_pattern <SEP> mitigated\_by\_course\_of\_action </PATH>} similar to~\cite{rog}). We used the paths to build \textit{Chain-of-Thought (CoT)} explanations, first analyzing the question and identifying the possible starting node (that at dataset building phase corresponds to the entity filling the placeholder), than describing
% that describe 
the planner’s reasoning step by step (\textit{“Step 1: Follow the \texttt{uses\_attack\_pattern} relation from the malware to the attack pattern. Step 2: ...”}). 
% The idea is that the planner should first identify the starting node (that at building phase corresponds to the entity filling the placeholder) within the question and then follow the sequence of relations to reach the final entities requested by the question. 
At the end, the reasoning process is summarized as a relational trace in the format: \texttt{<PATH> relation\_1 <SEP> ... <SEP> relation\_n </PATH>}. 

In some cases, a query does not specify a clear starting node, for example, \textit{“List all ransomware detection strategies”}, where every ransomware instance serves as a starting point. To handle such cases, TITAN introduces four operators: \texttt{filter}, \texttt{select}, \texttt{exec\_common}, and \texttt{exec\_difference}. The \texttt{filter} operator limits the search to entities that meet certain conditions, helping the reasoning process to focus on relevant subsets (e.g., techniques related to DDoS or SQL Injection). For the previous question, the planner should identify the starting \textit{ransomware} with the following relationships: \texttt{is\_malware\_type} (omitted in Figure~\ref{fig:ontology} for clarity, but used to retrieve nodes of a specific type) and \texttt{filter ransomware}. The \texttt{select} operator divides reasoning into parallel branches for different entities (in fact, there are question templates with multiple placeholders), each one following the same consecutive reasoning paths. 
\texttt{exec\_common} (to find shared nodes) and \texttt{exec\_difference} (to find disjunctive nodes) are used to compare results at the end, once all the paths in each branch, created by \texttt{select}, have been traversed. The generated dataset contains 74{,}258 training and 13{,}951 test samples, where the test set contains nodes and questions rearranged into unseen combinations. Figure \ref{fig:titan_examples_blocks} presents example questions with their corresponding paths, while Table \ref{tab:fine_profile_transposed} classifies samples by path length (L1–L4+) and the presence of operators, distinguishing single-hop (L1) from multi-hop (L2–L4+) reasoning.
\begin{table}[t]
\centering
\small
\setlength{\tabcolsep}{4pt}
\begin{tabular}{@{}lrrrr|rrrr@{}}
\toprule
& \multicolumn{4}{c}{\textit{Path length}} & \multicolumn{4}{c}{\textit{Operators}} \\
\cmidrule(lr){2-5}\cmidrule(lr){6-9}
\textbf{Split} & L1 & L2 & L3 & L4+ & \texttt{filter} & \texttt{select} & \texttt{exec\_common} & \texttt{exec\_diff} \\
\midrule
Train & 41{,}326 & 24{,}618 & 5{,}892 & 2{,}422 & 21{,}222 & 1{,}274 & 1{,}300 & 276 \\
Test  & 7{,}714  & 4{,}663  & 1{,}118 & 456   & 4{,}002  & 244   & 241   & 57 \\
\bottomrule
\end{tabular}
\caption{\footnotesize Dataset profile (Train $\xrightarrow{}$ 74{,}258, Test $\xrightarrow{}$ 13{,}951, Overall $\xrightarrow{}$ 88{,}209).}
\label{tab:fine_profile_transposed}
\end{table}

\section{Experiments}
\label{sec:experiments}
\begin{figure*}[t]
\centering

% --- Subfig A: FIGURE (plot: due pannelli verticali) ---
\begin{minipage}[c]{0.49\textwidth} % [c] = vertical center
  \centering
  \vspace{0pt} % allinea top interno
\begin{tikzpicture}[baseline=(current bounding box.north)]
\begin{groupplot}[
  group style={group size=1 by 2, vertical sep=14pt},
  width=\linewidth,
  height=0.55\linewidth,
  ymin=0, ymax=1,
  % ymajorgrids,
  % grid style={dashed,gray!20},
  every axis plot/.append style={thick},
  every node near coord/.append style={
    /pgf/number format/fixed,
    /pgf/number format/precision=2
  },
]

% ---------- AXIS 1: length ----------
\nextgroupplot[
  title={\textbf{Path Accuracy}},
  ylabel={},
  symbolic x coords={L1,L2,L3,{L4+}},
  xtick=data,
  legend to name=paLegend,
  legend columns=2,
  legend style={draw=none, fill=none, /tikz/every even column/.style={column sep=8pt}},
]
\addplot+[mark=*, solid, nodes near coords,
          nodes near coords style={font=\scriptsize, yshift=0.5pt}]
  coordinates {(L1,0.769) (L2,0.662) (L3,0.669) ({L4+},0.566)};
\addlegendentry{CoT}
\addplot+[mark=square*, solid, nodes near coords,
          nodes near coords style={font=\scriptsize, yshift=-12pt}]
  coordinates {(L1,0.575) (L2,0.480) (L3,0.553) ({L4+},0.368)};
\addlegendentry{NoCoT}

% --- label interna "length" in alto a sinistra del riquadro
\node[anchor=west, font=\footnotesize\bfseries,
      fill=white, fill opacity=0.7, text opacity=1, inner sep=2pt, rounded corners=1pt]
      at (rel axis cs:0.02,0.15) {length};

% ---------- AXIS 2: operators ----------
\nextgroupplot[
  ylabel={},
  symbolic x coords={filter,select,{diff.},{com.}},
  xtick=data,
  xticklabel style={rotate=25, anchor=east},
]
\addplot+[mark=*, solid, nodes near coords,
          nodes near coords style={font=\scriptsize, yshift=0.5pt}]
  coordinates {(filter,0.798) (select,0.676) ({diff.},0.456) ({com.},0.378)};
\addplot+[mark=square*, solid, nodes near coords,
          nodes near coords style={font=\scriptsize, yshift=-12pt}]
  coordinates {(filter,0.633) (select,0.561) ({diff.},0.421) ({com.},0.232)};

% --- label interna "operators" in alto a sinistra del riquadro
\node[anchor=west, font=\footnotesize\bfseries,
      fill=white, fill opacity=0.7, text opacity=1, inner sep=2pt, rounded corners=1pt]
      at (rel axis cs:0.02,0.15) {operators};

\end{groupplot}
\end{tikzpicture}

% % \begin{center}\pgfplotslegendfromname{paLegend}\end{center}
% \pgfplotslegendfromname{paLegend}\par\vspace{-0.6\baselineskip} % (opz.) compensa lo spazio
\begin{center}
\vspace{-15pt}\pgfplotslegendfromname{paLegend}\vspace{-15pt}
\end{center}
  \subcaption{CoT vs. NoCoT: Path Acc.}
  \label{fig:cot-nocot-sub}
\end{minipage}
\hfill
\begin{minipage}[c]{0.50\textwidth} % [c] = vertical center
\centering
% \vspace{0pt} % <-- forza l'allineamento in alto
\footnotesize
\setlength{\tabcolsep}{4pt}
\renewcommand{\arraystretch}{0.9}
\begin{tabular}{l|S S S S}
\toprule
\textbf{Bucket} & {\textbf{R-L}} & {\textbf{R-1}} & {\textbf{BLEU}} & {\textbf{BERT}} \\
\midrule
% --- length ---
L1   & 0.953 & 0.956 & 0.926 & 0.977 \\
L2   & 0.942 & 0.948 & 0.905 & 0.974 \\
L3   & 0.944 & 0.951 & 0.905 & 0.977 \\
L4+  & 0.919 & 0.933 & 0.864 & 0.967 \\
\cmidrule(lr){1-5}
% --- operators ---
\texttt{filter}            & 0.962 & 0.969 & 0.938 & 0.983 \\
\texttt{select}            & 0.955 & 0.962 & 0.922 & 0.979 \\
\texttt{diff.}  & 0.944 & 0.946 & 0.907 & 0.973 \\
\texttt{comm.}      & 0.830 & 0.864 & 0.752 & 0.932 \\
\bottomrule
\end{tabular}
  \subcaption{Reasoning metrics (CoT).}
  \label{tab:reasoning-metrics-sub}
\end{minipage}

\caption{(left)\textit{ CoT }vs. \textit{NoCoT}: Path Accuracy; (right) CoT reasoning metrics. Using diff $\xrightarrow{}$ \texttt{exec\_diff}, comm. $\xrightarrow{}$ \texttt{exec\_common}.}
\label{fig:cot-nocot-with-table}
\end{figure*}
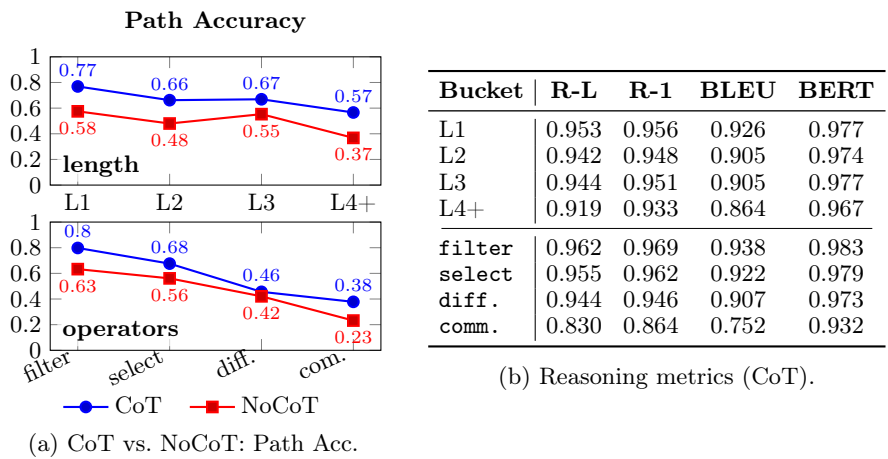

We evaluate path planner model using the test split in terms of \textit{Path Accuracy} and \textit{Reasoning abilities}. In particular, we also want to assess the impact of explicit \textit{Chain-of-Thought} (CoT) reasoning on the generation of executable relational paths in the TITAN ontology.  
We used Phi-3.5\footnote{\url{microsoft/Phi-3.5-mini-instruct}} as path planner, trained both with \textit{CoT} resoning and a \textit{NoCoT} variant to directly predict the path without intermediate reasoning. Note that, in the latter case, the graph executor is the one that finds the starting node, if present in the query, comparing all the words in the query with the list of node names inside the graph. We report results by \textbf{path length} (L1–L4+) and by \textbf{operator type} (\texttt{filter}, \texttt{select}, \texttt{exec\_difference}, \texttt{exec\_common}).  
\textit{Path Accuracy} is computed as the exact match (EM)~\cite{rajpurkar2016squad} between the predicted and reference paths.  
For CoT, we also evaluate the quality of the generated reasoning over the Ground Truth using ROUGE~\cite{lin2004rouge,lin2003automatic}, BLEU~\cite{papineni2002bleu}, and BERTScore~\cite{zhang2019bertscore}. As shown in Figure~\ref{fig:cot-nocot-sub}, the CoT model consistently outperforms the NoCoT baseline across all buckets.  
The largest improvements appear on longer paths (L4+) and filter-based operators, confirming that explicit reasoning helps maintain structural coherence in multi-hop queries.  
Conversely, set-based operators (\texttt{exec\_difference}, \texttt{exec\_common}) remain the most challenging due to their compositional nature. Reasoning metrics in Table~\ref{tab:reasoning-metrics-sub} show that CoT explanations maintain high linguistic and semantic alignment with references (ROUGE-L and BERTScore above 0.94 on average), with a mild drop for longer or more complex queries. Overall, explicit reasoning improves both the correctness and interpretability of the answers.

\section{Conclusions and Future Work}
\label{sec:conclusions}
In this work, we introduced \textit{TITAN}, a framework that autonomously traverses the TITAN Knowledge Graph (KG) to answer Cyber Threat Intelligence (CTI) questions.
TITAN integrates a \textit{path planner} LLM inspired by~\cite{rog}, which predicts relational paths and starting entities from natural-language queries, with a \textit{graph executor} that retrieves and validates the corresponding information from the KG. Future work will focus on enriching the TITAN KG with real malware reports and extending its ontology to support more expressive retrieval and analytical capabilities. We also plan to evaluate TITAN against existing CTI benchmarks and refine the planner’s reasoning through Reinforcement Learning–based alignment. Looking ahead, TITAN moves toward more interpretable systems able to reason over complex cyber knowledge. 

% By combining natural-language understanding with graph-based reasoning, TITAN aims to act as an autonomous CTI assistant that can retrieve, connect, and explain threat information in real time.

\bibliographystyle{splncs04}
\bibliography{bibliography}

\begin{thebibliography}{10}
\providecommand{\url}[1]{\texttt{#1}}
\providecommand{\urlprefix}{URL }
\providecommand{\doi}[1]{https://doi.org/#1}

\bibitem{alajramy2025device}
Alajramy, L., Simoni, M., Rasori, M., Saracino, A., Mori, P.: On-device derivation of iot usage control policies: Automating u-xacml policy generation from natural language with llms in smart homes environments. Future Generation Computer Systems p. 108067 (2025)

\bibitem{survey-cyber-threat-detection}
Chen, Y., Cui, M., Wang, D., Cao, Y., Yang, P., Jiang, B., Lu, Z., Liu, B.: A survey of large language models for cyber threat detection. Comput. Secur.  \textbf{145},  104016 (2024)

\bibitem{fieblinger2024actionable}
Fieblinger, R., Alam, M.T., Rastogi, N.: Actionable cyber threat intelligence using knowledge graphs and large language models. In: 2024 IEEE European symposium on security and privacy workshops (EuroS\&PW). pp. 100--111. IEEE (2024)

\bibitem{lin2004rouge}
Lin, C.Y.: Rouge: A package for automatic evaluation of summaries. In: Text summarization branches out. pp. 74--81 (2004)

\bibitem{lin2003automatic}
Lin, C.Y., Hovy, E.: Automatic evaluation of summaries using n-gram co-occurrence statistics. In: Proceedings of the 2003 human language technology conference of the North American chapter of the association for computational linguistics. pp. 150--157 (2003)

\bibitem{GRaph-LLM}
Liu, H., Wang, S., Zhu, Y., Dong, Y., Li, J.: Knowledge graph-enhanced large language models via path selection. In: Ku, L., Martins, A., Srikumar, V. (eds.) Findings of the Association for Computational Linguistics, {ACL} 2024, Bangkok, Thailand and virtual meeting, August 11-16, 2024. pp. 6311--6321. Association for Computational Linguistics (2024)

\bibitem{liu2020stix}
Liu, Z., Sun, Z., Chen, J., Zhou, Y., Yang, T., Yang, H., Liu, J.: Stix-based network security knowledge graph ontology modeling method. In: Proceedings of the 2020 3rd International Conference on Geoinformatics and Data Analysis. pp. 152--157 (2020)

\bibitem{rog}
Luo, L., Li, Y., Haffari, G., Pan, S.: Reasoning on graphs: Faithful and interpretable large language model reasoning. In: The Twelfth International Conference on Learning Representations, {ICLR} 2024, Vienna, Austria, May 7-11, 2024. OpenReview.net (2024)

\bibitem{noel2016cygraph}
Noel, S., Harley, E., Tam, K.H., Limiero, M., Share, M.: Cygraph: graph-based analytics and visualization for cybersecurity. In: Handbook of statistics, vol.~35, pp. 117--167. Elsevier (2016)

\bibitem{papineni2002bleu}
Papineni, K., Roukos, S., Ward, T., Zhu, W.J.: Bleu: a method for automatic evaluation of machine translation. In: Proceedings of the 40th annual meeting of the Association for Computational Linguistics. pp. 311--318 (2002)

\bibitem{rajpurkar2016squad}
Rajpurkar, P., Zhang, J., Lopyrev, K., Liang, P.: Squad: 100,000+ questions for machine comprehension of text. arXiv preprint arXiv:1606.05250  (2016)

\bibitem{XAI-for-cybersecurity}
Sarker, I.H., Janicke, H., Mohsin, A., Gill, A., Maglaras, L.: Explainable {AI} for cybersecurity automation, intelligence and trustworthiness in digital twin: Methods, taxonomy, challenges and prospects. {ICT} Express  \textbf{10}(4),  935--958 (2024)

\bibitem{MATRIX}
Simoni, M., Saracino, A.: {MATRIX:} {A} comprehensive graph-based framework for malware analysis and threat research. In: di~Vimercati, S.D.C., Samarati, P. (eds.) Proceedings of the 22nd International Conference on Security and Cryptography, {SECRYPT} 2025, Bilbao, Spain, June 11-13, 2025. pp. 495--502. {SCITEPRESS} (2025). \doi{10.5220/0013629300003979}, \url{https://doi.org/10.5220/0013629300003979}

\bibitem{cyb_with_llm_rag}
Simoni, M., Saracino, A.: Cybersecurity with llms and rags: Challenges and innovations. In: Alrabaee, S., Choo, K.K.R., Damiani, E., Deng, R.H. (eds.) Security and Privacy in Communication Networks. pp. 169--183. Springer Nature Switzerland, Cham (2026)

\bibitem{simoni2025morse}
Simoni, M., Saracino, A., P, V., Conti, M.: Morse: Bridging the gap in cybersecurity expertise with retrieval augmented generation. In: Proceedings of the 40th ACM/SIGAPP Symposium on Applied Computing. pp. 1213--1222 (2025)

\bibitem{strom2018mitre}
Strom, B.E., Applebaum, A., Miller, D.P., Nickels, K.C., Pennington, A.G., Thomas, C.B.: Mitre att\&ck: Design and philosophy. In: Technical report. The MITRE Corporation (2018)

\bibitem{IBM-KG-path}
Wang, S., Lin, J., Guo, X., Shun, J., Li, J., Zhu, Y.: Reasoning of large language models over knowledge graphs with super-relations. In: The Thirteenth International Conference on Learning Representations, {ICLR} 2025, Singapore, April 24-28, 2025. OpenReview.net (2025)

\bibitem{wei2022chain}
Wei, J., Wang, X., Schuurmans, D., Bosma, M., Xia, F., Chi, E., Le, Q.V., Zhou, D., et~al.: Chain-of-thought prompting elicits reasoning in large language models. Advances in neural information processing systems  \textbf{35},  24824--24837 (2022)

\bibitem{uncovering-attacks-with-KG}
Xiang, X., Ma, C., Zeng, L., Feng, W., Xie, Y., Gu, Z.: Uncovering multi-step attacks with threat knowledge graph reasoning. Secur. Saf.  \textbf{4},  2024019 (2025). \doi{10.1051/SANDS/2024019}, \url{https://doi.org/10.1051/sands/2024019}

\bibitem{zhang2019bertscore}
Zhang, T., Kishore, V., Wu, F., Weinberger, K.Q., Artzi, Y.: Bertscore: Evaluating text generation with bert. arXiv preprint arXiv:1904.09675  (2019)

\end{thebibliography}

\end{document}